\documentclass{article}

\usepackage{arxiv}

\usepackage[utf8]{inputenc} 
\usepackage[T1]{fontenc}    
\usepackage[colorlinks=true,allcolors=black]{hyperref}       
\usepackage{url}            
\usepackage{booktabs}       
\usepackage{amsfonts}       
\usepackage{nicefrac}       
\usepackage{microtype}      
\usepackage{lipsum}		
\usepackage{graphicx}
\usepackage[numbers]{natbib}
\usepackage{doi}
\usepackage{pifont}
\usepackage{xcolor}
\usepackage{amsmath}
\usepackage{fancyhdr}

\title{Exploring Knowledge Tracing in Tutor-Student Dialogues using LLMs}

\author{Alexander Scarlatos\\
University of Massachusetts Amherst\\
ajscarlatos@cs.umass.edu\\
\And
Ryan S. Baker\\
University of Pennsylvania\\
rybaker@upenn.edu
\And
Andrew Lan\\
University of Massachusetts Amherst\\
andrewlan@cs.umass.edu
}

\date{}

\renewcommand{\headeright}{}
\renewcommand{\undertitle}{Published in LAK25:
The 15th International Learning Analytics and Knowledge Conference}

\begin{document}
\maketitle

\begin{abstract}
  Recent advances in large language models (LLMs) have led to the development of artificial intelligence (AI)-powered tutoring chatbots, showing promise in providing broad access to high-quality personalized education. Existing works have studied how to make LLMs follow tutoring principles, but have not studied broader uses of LLMs for supporting tutoring. Up until now, tracing student knowledge and analyzing misconceptions has been difficult and time-consuming to implement for open-ended dialogue tutoring. In this work, we investigate whether LLMs can be supportive of this task: we first use LLM prompting methods to identify the knowledge components/skills involved in each dialogue turn, i.e., a tutor utterance posing a task or a student utterance that responds to it. We also evaluate whether the student responds correctly to the tutor and verify the LLM's accuracy using human expert annotations. We then apply a range of knowledge tracing (KT) methods on the resulting labeled data to track student knowledge levels over an entire dialogue. We conduct experiments on two tutoring dialogue datasets, and show that a novel yet simple LLM-based method, LLMKT, significantly outperforms existing KT methods in predicting student response correctness in dialogues. We perform extensive qualitative analyses to highlight the challenges in dialogueKT and outline multiple avenues for future work.
\end{abstract}


\section{Introduction}

Tutoring, often in the form of natural-language dialogues between tutors and students, has proven to effectively improve student learning outcomes, in both human tutoring \cite{metaat} and dialogue-based intelligent tutoring systems \cite{nye2014autotutor}. Recent advances in generative artificial intelligence (AI), especially large language models (LLMs), have led to the development of generative AI-powered tutoring chatbots, such as Khan Academy's Khanmigo \cite{khanmigo} and Carnegie Learning's LiveHint \cite{livehint}. By wrapping LLMs around learning content such as textbooks and practice questions, these automated chatbots can interact with students in real time, potentially reaching a larger number of students than human tutors, while greatly reducing authoring efforts relative to previous-generation dialogue tutors \cite{jordan2001tools}. 

To use LLMs as tutors, it is important to i) ensure they use effective pedagogical strategies and ii) control the safety of their output. Most existing LLM-powered tutoring chatbots are developed by either fine-tuning on actual tutor utterances in tutoring dialogue transcripts or prompt engineering. Along these lines, plenty of works over the last two years have studied how to make LLMs follow effective tutoring principles. The work in \cite{wang2024bridging} studies the latent decision-making process behind human tutor utterances and shows that adopting decisions by human experts can benefit LLMs in tutoring. The work in \cite{macina-etal-2023-mathdial} identifies tutor moves when they interact with an LLM-powered simulated student agent and shows that move annotation data can make LLMs become better tutors. The work in \cite{learnlm} generates large amounts of synthetic tutoring data via AI roleplay and shows that fine-tuning on this data, in addition to data collected from human tutors and students, makes LLMs more effective in pedagogy. The work in \cite{demszky2021measuring} proposes a way to measure tutor uptake, i.e., acknowledging what students have said, which is another beneficial strategy for LLMs to adopt. The work in \cite{prihar2023comparing} attempted to summarize tutor utterances into actionable feedback for students using LLMs, but found that the resulting messages were unable to outperform those written by real human teachers. 

One important observation is that existing works involving LLMs have almost exclusively studied the \emph{tutor turns} in dialogues rather than the \emph{student turns}. A possible reason is that student turns can be noisy in nature, partially due to students not fully mastering the required knowledge, resulting in utterances that contain errors, which are harder to analyze than tutor utterances. Take Google's LearnLM \cite{learnlm} for example: student dialogue turns in their AI roleplay-based synthetic training data are generated by prompting an LLM to make an error that a real student would make. This error information is then fed to the tutor LLM to help the tutor model identify the error and provide corresponding feedback. However, there is a key limitation to this approach: assuming LLMs can behave like real students may not be reliable. As recent work \cite{fernandez2024divert} shows, LLMs are ineffective at anticipating, parsing, and following flawed reasoning. More importantly, real open-ended student discourse is highly valuable since it may indicate flaws in student knowledge and even reveal specific misconceptions. Therefore, analyzing student dialogue turns can be a form of formative assessment to help tutors and AI-powered chatbots understand student progress and provide more targeted feedback. It can also potentially improve simulated student agents \cite{kaser2024simulated} to make them better at reflecting real student behavior, which can, in turn, improve the training of tutor chatbots through reinforcement learning \cite{rafferty2016faster,he2021quizzing}.

\subsection{Contributions}
In this work, we propose dialogue knowledge tracing (dialogueKT), a novel task, which analyzes student discourse within the knowledge tracing (KT) framework \cite{pelanek2017bayesian}.
We enable KT on dialogues via a two-step process: we i) use proprietary LLMs, specifically GPT-4o \cite{gpt-4o}, to annotate dialogues with student response correctness and knowledge component (KC) tags using Common Core standards, and ii) train KT models on the annotated data to estimate student knowledge at the dialogue-turn level. We introduce a novel LLM-based KT method, LLMKT, that leverages the textual content in dialogues, by fine-tuning the open-source Llama 3 LLM \cite{llama31} on the KT objective. 
Using two existing tutor-student math dialogue datasets, we show that i) our new methods significantly outperform existing KT methods, especially when little training data is available, and ii) all KT methods can accurately predict student correctness given enough training data. We also use qualitative analysis and learning curve visualizations to show that LLMKT learns meaningful knowledge state estimates. Finally, we show that GPT-4o's dialogue annotations are accurate, according to human experts. Overall, by assessing student performance in more detail, we can make it possible to apply KT to LLM-human tutoring dialogues, enabling mastery learning and various other pedagogies and forms of detection afforded by KT \cite{pelanek2017bayesian}.

We acknowledge many limitations of our work upfront since it is, to the best of our knowledge, the first at attempting to perform KT in open-ended tutoring dialogues. First, a ``time step'' in our KT setup corresponds to a single student turn in the dialogue, which is highly granular; 
we cannot explore setting an entire dialogue as a time step since no existing open-source large-scale tutoring dialogue dataset links students across dialogues.
Second, our work is restricted to math since analyzing the KCs in math problems is challenging yet well-studied; it remains to be seen whether our approach of Common Core standard tagging generalizes to other subjects. Third, due to the smaller scale of tutoring dialogue datasets compared to those typically used in standard KT works, our methods need to be tested on larger-scale data. One dataset we use in our experiments contains partial dialogues that are cut short for LLM evaluation purposes \cite{miller_dicerbo_2024}, which further adds to this issue. Therefore, we publicly release our code\footnote{\url{https://github.com/umass-ml4ed/dialogue-kt}} so that researchers who own a large amount of dialogue data can experiment on their proprietary data.

\section{Related Work}

\subsection{Knowledge Tracing and Knowledge Component Labeling}

Knowledge tracing (KT) \cite{pelanek2017bayesian} is a well-studied task in the student modeling literature. It breaks down student learning into a series of time steps practicing certain KCs and uses the correctness of each response step to track student knowledge. Classic Bayesian KT methods use a latent binary-valued variable to represent whether a student masters a KC or not. Factor analysis-based methods \cite{das3h,pfa} use features in addition to latent ability parameters in a logistic regression setting to predict student response correctness.
Since neural networks became popular, plenty of deep learning-based KT methods were developed.
These models have limited interpretability since student knowledge is represented as hidden states in complex, deep neural networks. Most of these methods use long short-term memory networks \citep{dkt} or variants \cite{saint+}, coupled with with memory augmentation \cite{dkvmn}, graph neural networks \cite{gikt}, attention networks \cite{akt,sakt}, or pre-trained word embeddings in question text \cite{eernna}.
Many recent works have attempted to use LLMs in some ways for KT. Some use the neural network architecture of LLMs and train on student response data \cite{zhan2024knowledge}. Others use LLMs to represent questions and responses in certain domains such as math \cite{jung2024clstcoldstartmitigationknowledge}, language learning \cite{srivastava2021question,cui2023adaptive}, and computer science education \cite{lee2024languagemodelknowledgetracing,okt}.
More recently, \cite{li2024explainable} attempted prompting-based KT that directly asks LLMs to represent questions and student knowledge states as text.
Despite some success in question response modeling, it remains to be seen whether these models can be useful in text-heavy KT situations such as dialogues. 

There exist many prior works on automated KC tagging and discovery, including the Q-matrix method \cite{barnes2005q}, which assigns KCs to questions using student response data. More recently, methods such as \cite{kc-finder} also learn latent KCs by training deep learning models on KT data and enforcing priors from pedagogical theory. While these methods effectively learn new KCs, they do not produce textual descriptions of KCs, making them difficult to interpret. Several recent works have used LLMs to assign expert-written KCs to math problems via prompting \cite{atc,yang2024content}, fine-tuning \cite{shen2021classifying}, and reinforcement learning \cite{li2024knowledgetaggingmathquestions}. However, none have attempted assigning KCs in open-ended dialogues. A prior work has attempted to tag KCs in dialogue turns \cite{croteau2004algebra}, although using a manual approach that cannot be scaled.

\subsection{Dialogue Analysis}

Existing works on dialogue analysis using LLMs have primarily focused on analyzing tutor behavior. The works in \cite{ncte,suresh-etal-2022-talkmoves} automatically classify the ``talk moves'' that tutors and students make via fine-tuned LLMs, and the work in \cite{abdelshiheed2024aligning} uses talk moves in dialogues to predict student performance in post-tests.
In pre-LLM dialogue tutors, AutoTutor used latent semantic analysis to infer whether a student response was correct, and when incorrect, whether misconceptions were involved \cite{cai2018impact}. Why2-Atlas used deep syntactic analysis to convert a student response to a proof, analyzed the proof for errors represented as lack of knowledge \cite{vanlehn2002architecture}, and later used Bayesian networks \cite{makatchev2007combining}. The Watson Tutor used ensembles of different classifiers on a range of textual features for correctness evaluation \cite{chang2018dialogue} and Glicko-2 \cite{glickman2012example} for KT.

\section{Knowledge Tracing in Dialogues}

\begin{figure*}[t]
    \centering
    \includegraphics[width=.85\linewidth]{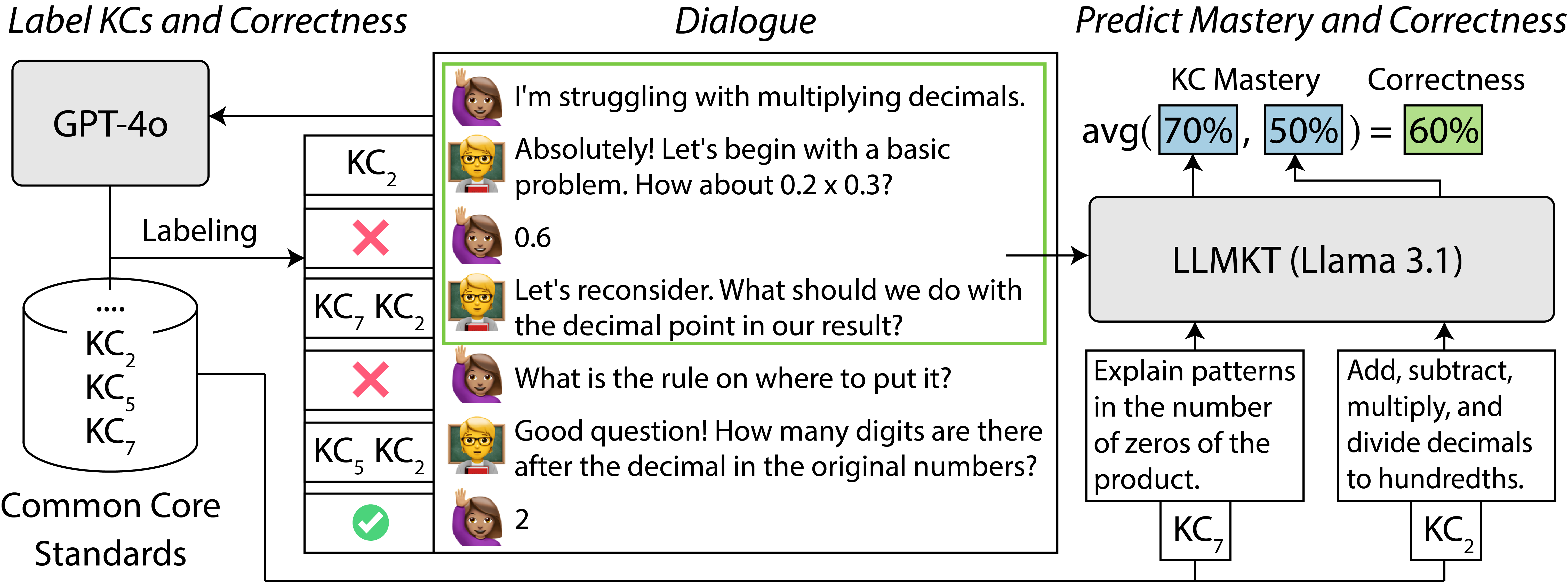}
    \caption{Overview of the dialogueKT framework and LLMKT. For annotation, we provide a full dialogue to GPT-4o, which generates correctness labels and retrieves Common Core standards as KCs for each turn. For KT, LLMKT takes a dialogue up to the target turn and its KCs to predict mastery on each KC and response correctness.}
    \label{fig:arch}
\end{figure*}

We now formulate the dialogueKT framework. 
Our goal is to develop methods that can i) estimate student knowledge on KCs involved in each dialogue turn and ii) use these estimates to predict the correctness of student dialogue turns. Intuitively, we can view tutor-student dialogues as sequences of (formative) assessments. Visualized in Figure~\ref{fig:arch}, some tutor turns can be viewed as posing a task that is related to some learning objectives and one or more KCs. They can reflect different tutor moves: asking a clarification question, identifying a student error, providing actionable feedback, prompting the student to think in a Socratic way, or offering encouragement \cite{suresh2022fine,demszky2021measuring}. Student turns, on the other hand, contain correct or incorrect responses to tasks posed by the tutor, which reflect their thinking process and offer insights into their knowledge state. 
A pair of tutor-student dialogue turns corresponds to a time step in KT. Therefore, we split the dialogueKT framework into three stages: i) identify the KCs involved in each student dialogue turn, ii) classify whether the student response is correct or incorrect, and iii) apply new and existing KT methods on the resulting data.

Formally, we define each dialogue $d$ to be a sequence of alternating tutor turns, $t$, and student turns, $s$, i.e., $d=(s_0,t_1,s_1,\ldots,t_M,s_M)$, where $M$ is the total number of turn pairs for that dialogue and $s_0$ is present if the student initiates the dialogue. Each relevant turn pair is assigned a correctness label $y_j \in \{0, 1\}$, where $j \in \{1,\ldots,M\}$ is the turn pair index; $y_j=1$ means that $s_j$ is a correct response to $t_j$, and $y_j=0$ means that it is incorrect. In practice, we also use $y_j=\text{na}$ in cases where correctness is not well-defined for $t_j$, e.g., the turn is off-topic or the tutor is offering emotional support. Each tutor turn is also associated with a set of KCs, $\mathcal{C}_j=\{c_{j1},\ldots,c_{jK}\}$, where $K$ is the number of KCs for that turn. These KCs are related to learning objectives outlined by the tutor, which lists knowledge elements that the student must posses to respond correctly. 
In this work, we use the Common Core math standards \cite{commoncore} as the KCs since they are i) standard and widely adopted and ii) textual in nature and are interpretable to both humans and LLMs. 
For irrelevant dialogue turns that we do not consider, $\mathcal{C}_j=\emptyset$.

In a dialogue turn, we use $z_{jk} \in \{0,1\}$ to denote the student's unobserved, binary-valued latent mastery status of the KC, $c_{jk}$, that is involved in the current step; $z_{jk} = 1$ means that the student masters KC $k$ in dialogue turn $j$. In our framework, the student’s mastery of all KCs involved in a turn determines whether they will respond correctly to the task posed by the tutor. Formally, the probability of a correct student response at turn $j$ can be written as $P(y_j = 1 | z_{j1}, \ldots, z_{jK})$. The latent KC mastery depends on the entire dialogue history, i.e., $P(z_{jk}=1) = \phi(c_{jk}, t_{\le j}, s_{<j}, y_{<j}, \mathcal{C}_{< j})$, where $\phi$ returns the probability of the student mastering KC $c_{jk}$, and subscripts $<j$ and $\le j$ correspond to all past student and (including the current) tutor dialogue turns, respectively. In the next section, we detail how we parameterize and train this function in practice.

\section{Methodology}

We now detail our two primary technical contributions. First, we detail how to prompt pre-trained LLMs to automatically annotate dialogue turns to collect the KC and response correctness labels needed for dialogue knowledge tracing. Second, we propose new KT methods for the dialogueKT task that incorporate the rich textual information in tutor-student dialogues. We visualize the dialogueKT framework and the LLMKT method in Figure~\ref{fig:arch}.

\subsection{Automated Dialogue Annotation}

In order to perform knowledge tracing on dialogues, each student conversation turn must first be annotated with two labels: i) the correctness of the student response, and ii) a list of KCs involved in the turn. While it is possible for human experts to provide these annotations, doing so would be incredibly time consuming and difficult to scale. Following recent works that have had success using LLMs to annotate data in educational tasks \cite{wang2024bridging, scarlatos2024feedback}, we use a state-of-the-art LLM, GPT-4o \cite{gpt-4o}, to perform the annotation via simple, zero-shot chain-of-thought (CoT) prompting\footnote{We release our prompts in an online supplementary material: \url{https://osf.io/873ms?view_only=f32472a6560649e692a4e2ba603ef52c}}. We conduct the two annotation tasks separately since we find that doing so leads to improved accuracy compared to combining them. 
In Section~\ref{sec:annotation-experiments}, we conduct a human evaluation with math teachers to show that the GPT-provided annotations are mostly accurate and aligned with human judgment. 

For correctness annotation, we provide the full dialogue as input to the LLM and instruct it to first summarize each turn in the dialogue in a CoT way, before labeling the correctness of each turn. We specifically instruct the LLM to ``summarize'' to leverage the fact it is pre-trained extensively on summarization tasks; this approach is inspired by recent work \cite{ma2024debugging} that has shown that such an approach is effective for tasks such as code editing.

For KC annotation, we adapt the recursive tagging algorithm from \cite{atc}, which tags math word problems with Common Core math standards. The algorithm uses hierarchically organized standards from the Achieve the Core\footnote{\url{https://achievethecore.org/}} (ATC) coherence map, where standards are grouped into mid-level ``clusters'', which are further grouped into high-level ``domains''. To assign standards to a dialogue, we first provide the LLM with the full dialogue and the list of all 11 domains, and ask it to summarize the dialogue before selecting all domains that are relevant to the learning objectives. We then take the union of all clusters that are children of the selected domains and prompt the LLM again, this time asking it to summarize at the turn level but still select relevant clusters at the dialogue level. Finally, we take the union of all standards that are children of the selected clusters and prompt the LLM a final time, having it summarize each turn and assign standards at the turn level. We treat each Common Core math standard as a KC.

We note that by annotating a full dialogue with a single prompt, KT in real-time becomes impossible since information from future turns is available when annotating any given turn. This limitation could be alleviated in practice by annotating dialogues one turn at a time; however, such an approach is costly and we leave that for future work.

\subsection{LLMKT}

Another feature that makes dialogueKT different than the regular KT task is that most dialogue turns (at least 58\%) are associated with more than one KC; see Table~\ref{tab:data-stats} for details. In contrast, most existing KT methods are developed for questions on a single KC; solutions that can deal with the occasional multi-KC question include i) converting a single question-response pair with multiple KCs to multiple pairs for different ``questions'', each covering a single KC, and ii) treating a unique combination of multiple KCs as a new KC. However, due to the prevalence of dialogue turns tagged with multiple KCs, we found that these solutions are not applicable to our setting.
Therefore, we consider the student's likelihood of responding correctly to a tutor-posed task to be their average mastery of all KCs involved in the turn. Empirically, we found that averaging over KC masteries performed better than taking a product over them, consistent with findings in prior work \cite{maier2021challenges}. Formally, we model the probability of a correct student response at a turn as
\begin{align}
P(y_{j}=1) = \textstyle \frac{1}{K} \sum_{k=1}^K P(z_{jk}=1). \label{eq:independence}
\end{align}
In other words, we use a compensatory model rather than a conjunctive model \cite{maier2021challenges}. We now detail our proposed LLM-based KT method, which we refer to as LLMKT. One can, in theory, estimate a student knowledge state, $z_{jk}$, by simply providing the text of the dialogue up to $t_j$ to a pre-trained LLM and prompting it to predict whether the student masters the KC, $c_{jk}$, for each $k \in [1,K]$. However, since LLMs are not pre-trained on the specific KT task, we cannot expect them to do well in this ``zero-shot'' setting. Therefore, we adopt a fine-tuning approach using Llama-3.1-8B-Instruct, an open-source, decoder-only LLM, for the KT task. We model the estimated probability of a student mastery $\hat{z}_{jk}$ as
\begin{align}
    &\hat{z}_{jk} = P_\theta(z_{jk}=1) = \frac{\exp(v^\mathcal{T}_\theta(t_{\le j},s_{<j},c_{jk}))}{\textstyle \sum_{\text{tok} \in \{\mathcal{T}, \mathcal{F}\}} \exp(v^\text{tok}_\theta(t_{\le j},s_{<j},c_{jk}))}, \label{eq:softmax}
\end{align}
where $\theta$ represents the model parameters, and $v^\mathcal{T}_\theta(\cdot)$ and $v^\mathcal{F}_\theta(\cdot)$ return the output logits of the LLM for the ``True'' and ``False'' vocabulary tokens, respectively. In other words, with i) the entire conversation history and ii) a textual prompt instructing the LLM to estimate the student's current KC mastery level as input, we use an LLM to parameterize this estimate through two token generation probabilities. In practice, to improve model efficiency, we pack all KCs in $C_j$ into a single prompt; we use customized attention masks and position embeddings to ensure that KCs do not attend to each other and their mastery levels are estimated independently. We do not explicitly provide KCs or correctness labels for previous turns due to memory constraints, but observe that LLMs can infer them from the context. 

We fine-tune the LLM by maximizing the likelihood of all observed student response correctness labels, or equivalently, minimizing the binary cross entropy loss between the ground truth and predicted correctness values, $y_j$ and $\hat{y}_j$, as
\begin{align}
    &\hat{y}_j = \textstyle \frac{1}{K} \sum_{k=1}^K \hat{z}_{jk}\\ 
    &\mathcal{L}(d;\theta) = -\textstyle \sum_{j=1}^M y_j \log \hat{y}_j + (1 - y_j) \log (1 - \hat{y}_j). \label{eq:bce_loss}
\end{align}

While this method is relatively simple, it has many advantages. First, because we start with a powerful pre-trained LLM, LLMKT is able to leverage existing textual analysis capabilities that are not available in other KT architectures. Second, we do not add any new parameters to the pre-trained model, likely reducing the need for large amounts of training data. Third, since LLMKT operates on textual content rather than question/KC embeddings that need to be learned from scratch, it should be more generalizable to new KCs that are not seen during training. We note that LLMKT is similar to CLST \cite{jung2024clstcoldstartmitigationknowledge}, which trains an LLM to predict student correctness using a softmax over tokens (Eq.~\ref{eq:softmax}). However, unlike LLMKT, their method does not handle multiple KCs per item, which is critical to the dialogueKT task.

\subsection{DKT-Sem}

In addition to LLMKT, we experiment with an alternative, simpler KT method for dialogues, which can be viewed as a strong baseline. We slightly modify the deep KT (DKT) model \cite{dkt} to use semantic embeddings of the textual content in dialogues, resulting in a method we term DKT-Sem. We map the text embeddings of every dialogue turn to the latent knowledge space in DKT as input, and map the text embeddings of KCs to the latent knowledge space to predict KC masteries. We believe that this simple method can result in better generalization across KCs and dialogues compared to DKT, which learns question and KC representations from scratch. Formally, we model KC mastery probability as
\begin{align}
    &\mathbf{t}_j = \operatorname{S-BERT}(t_j), \quad \mathbf{s}_j = \operatorname{S-BERT}(s_j), \quad \mathbf{c}_j = \textstyle \frac{1}{K} \sum_{k=1}^K \operatorname{S-BERT}(c_{jk})\\
    &\mathbf{y}_j = \operatorname{emb}({y_j}), \quad \mathbf{x}_j = \tanh([\mathbf{t}_j; \mathbf{s}_j; \mathbf{c}_j]\mathbf{W}) \oplus \mathbf{y}_j\\
    &\mathbf{h}_j = \operatorname{LSTM}(\mathbf{x}_{\le j}), \quad \hat{\mathbf{z}}_{j+1} = \sigma(\mathbf{h}_{j}\mathbf{B}\mathbf{C}^T).
\end{align}
The input at each time step is created by first computing the sentence BERT (S-BERT) \cite{sbert} embeddings of the teacher turn, $\mathbf{t}_j$, student turn, $\mathbf{s}_j$, and element-wise average embeddings of the KCs, $\mathbf{c}_j$, respectively. Following DKT, response correctness is mapped to a learnable embedding, $\mathbf{y}_j$, which is then added element-wise to a transformed version (through $\mathbf{W}$ and $\tanh$) of the concatenated tutor turn/student turn/KC embeddings, resulting in the input vector $\mathbf{x}_j$. Finally, the vector of estimated KC masteries at the next turn, $\hat{\mathbf{z}}_{j+1}$, are computed via a bilinear projection, $\mathbf{B}$, between the LSTM hidden state, $\mathbf{h}_j$, and the matrix of all KC S-BERT embeddings, $\mathbf{C}$, followed by the sigmoid function. We train DKT-Sem using the same objective as LLMKT (Eq.~\ref{eq:bce_loss}). Due to the smaller model size, DKT-Sem is easier to train than LLMKT, especially when access to computational resources to support LLM fine-tuning is limited.

\section{Experiments}
\label{sec:expts}

\begin{table*}[t]
    \centering
    \begin{tabular}{lcccccccc}
        \toprule
        Dataset & Num.\ Dia.\ & Num.\ Labels & Correct & Num.\ KCs & Avg.\ KCs/Dia.\ & Avg.\ KCs/Turn & Turns w/ >1 KC \\
        \midrule
        CoMTA & 153 & 623 & 57.78\% & 164 & 3.20 & 1.94 & 58.75\% \\
        MathDial & 2,823 & 13,200 & 49.86\% & 145 & 4.16 & 2.32 & 82.89\% \\
        \bottomrule
    \end{tabular}
    \caption{Statistics for the CoMTA and MathDial datasets, from left to right: total number of dialogues, total number of turn pairs with correctness labels, percentage of turn pairs where the student response is correct, total number of unique KCs in the dataset, average number of unique KCs per dialogue, average number of KCs per turn pair, and percentage of turn pairs with more than one KC.}
    \label{tab:data-stats}
\end{table*}

\subsection{Datasets}

We experiment with two datasets, CoMTA and MathDial, to study the dialogueKT task and validate the proposed KT methods. 
We select CoMTA since it is recent and highly reflective of AI-student dialogues that are becoming prevalent, and we select MathDial since it is relatively large-scale. 
We acknowledge that there are many other datasets available, such as the ones in \cite{ncte,suresh2022fine,wang2024bridging} and ones not focusing on math education; our overall framework should be applicable to these as well. We show statistics for CoMTA and MathDial in Table~\ref{tab:data-stats}.

The CoMTA dataset \cite{miller_dicerbo_2024} contains 188 dialogues between human students and Khanmigo, Khan Academy's GPT-4-powered tutor. In most dialogues, students ask for help with a specific topic or problem(s) that they do not know how to solve, and work through one (or more) problems with the tutor throughout the dialogue. The tutor is primed with a series of customized prompts developed by Khan Academy to behave like real human tutors: to use the Socratic method, frequently asking questions at almost every dialogue turn, and to not directly reveal correct answers to students. Each dialogue centers around a math subject among Elementary, Algebra, Trigonometry, Geometry, and Calculus. However, since Common Core does not contain Calculus standards, we remove Calculus dialogues, resulting in 153 remaining dialogues. Since no train/test split is defined, we perform a 5-fold cross-validation in experiments.

The MathDial dataset \cite{macina-etal-2023-mathdial} contains 2,848 dialogues between students simulated by GPT-3.5 and crowd workers role-playing as human tutors. Each dialogue is in the context of an incorrect ``student'' solution to a math problem, and the tutor's goal is to help the student arrive at the correct solution by helping them identify and correct their errors. We use the original train/test split, containing 2,235/588 dialogues, respectively, after removing a total of 25 dialogues where GPT-4o's annotation of student response correctness and turn KCs failed.

\subsection{KT Methods}

In addition to LLMKT and DKT-Sem, we also evaluate the performance of a suite of widely-used KT methods on the dialogueKT task. Our goal is to test whether existing KT methods, developed for a slightly different use case, are applicable to our setting and how they compare to methods designed for dialogueKT.

We evaluate multiple popular deep learning-based  KT methods: \textbf{DKT} \cite{dkt}, \textbf{DKVMN} \cite{dkvmn}, \textbf{AKT} \cite{akt}, \textbf{SAINT} \cite{saint}, and \textbf{simpleKT} \cite{liu2023simplekt}, all sourced from the pyKT code repository \cite{liupykt2022}. There are two key differences between these existing KT methods and our proposed KT methods for dialogues: i) they learn new embeddings for questions, KCs, and response correctness without leveraging any textual information, and ii) they are designed to analyze questions that cover a single KC. Since most of the dialogue turns contain at least two KCs in the CoMTA and MathDial datasets, we convert each turn into $K$ ``pseudo-turns'' that are each associated with one of the KCs, $c_{jk}$; all such pseudo-turns are associated with the same correctness label, $y_j$, of that turn. This modification follows common practice in the KT literature \cite{akt,liu2023simplekt}. We predict a single $\hat{\mathbf{z}}_j$ vector for each turn pair, using the output of the most recent time step that does not take $y_j$ as input, to avoid leaking the correctness label across pseudo-turns. We note that predicting a vector instead of a single probability requires altering the size of the output projection for all methods except DKT. We train all methods using the same objective as LLMKT and DKT-Sem (Eq.~\ref{eq:bce_loss}). We also compare against Bayesian KT (\textbf{BKT}) \cite{kt}, a classical KT method that uses a hidden Markov model to learn interpretable, per-KC learning acquisition parameters and use them to predict changing mastery levels for a student. Since BKT is designed to model data where each question is only assigned a single KC, we similarly expand turns into ``pseudo-turns'' as described above.
Due to the rigidity of BKT, we use a slightly different setup compared to other methods. At train-time, we train on all pseudo-turn labels, rather than aggregating at the true turn level. At test-time, we take the average pseudo-turn prediction to be the true turn prediction.
In contrast to the other methods, this pseudo-turn level prediction setup does not leak correctness labels because information does not interact across KCs in BKT. We use pyBKT \cite{pybkt} to train and generate predictions.

\subsection{Metrics and Experimental Setup}

To quantitatively evaluate the effectiveness of KT methods in the dialogueKT task, we employ 3 widely used metrics: i) accuracy (\textbf{Acc.}), the portion of predicted labels that match the ground truth after rounding $\hat{y}_j$ to 0 or 1, ii) the area under the receiver operating characteristic curve (\textbf{AUC}), the most widely used metric in the KT literature as it accounts for class imbalance and all prediction thresholds, and iii) \textbf{F1}, the harmonic mean between binary precision and recall. We compute these metrics on all correctness labels except for the first label in each dialogue, since existing KT methods are unable to make educated estimates without prior labels in context.

We implement LLMKT using the Huggingface Transformers library \cite{wolf-etal-2020-transformers} and perform fine-tuning with LoRA \cite{hu2022lora} on NVIDIA RTX A6000 GPUs. We implement DKT-Sem using PyTorch and use the \texttt{all-mpnet-base-v2} S-BERT model to generate text embeddings. During training, we randomly take 20\% of the dialogues in the train set to use as a validation set and perform early stopping. We then evaluate all metrics on the test set, and repeat for all 5 folds on CoMTA.  

Before computing final results, we perform a grid search to find optimal hyperparameters for each method, keeping the checkpoint with the highest AUC on the validation set. For LLMKT, we perform the search on the CoMTA dataset and use the same parameters on MathDial to reduce training time, and for other methods we use the best hyperparameter values found for each dataset. For LLMKT, we search over learning rate $\in \{5\cdot10^{-5}, 1\cdot10^{-4}, 2\cdot10^{-4}, 3\cdot10^{-4}\}$ and LoRA's rank $r \in \{4, 8, 16, 32\}$, and fix LoRA's $\alpha$ to 16, batch size to 64 using gradient accumulation, gradient norm clipping to 1.0, and train for 5 epochs. For other KT methods, we search over learning rate $\in \{1\cdot10^{-4}, 2\cdot10^{-4}, 5\cdot10^{-4}, 1\cdot10^{-3}, 2\cdot10^{-3}, 5\cdot10^{-3}\}$ and embedding size $\in \{8, 16, 32, 64, 128, 256, 512\}$, and fix batch size to 64, train for 100 epochs, and set all other model-specific hyperparameters to the defaults in pyKT. For all methods, we train with the AdamW optimizer with a weight decay of $1\cdot10^{-2}$. We found the optimal learning rate and $r$ of LLMKT to be $2\cdot10^{-4}$ and 16, respectively; the optimal hyperparameters for other KT methods varied across methods and datasets, but tended to have smaller embedding sizes on CoMTA and larger ones on MathDial since the former is much smaller in scale compared to the latter.

\section{Experimental Results}

\subsection{Quantitative Results}

\begin{table*}[t]
    \centering
    \begin{tabular}{l|ccc|ccc}
        \toprule
        & \multicolumn{3}{c|}{CoMTA} & \multicolumn{3}{c}{MathDial}\\
        Method & Acc. & AUC & F1 & Acc. & AUC & F1 \\
        \midrule
        BKT & $51.02_{\pm 11.16}$ & $52.50_{\pm 10.61}$ & $53.41_{\pm 13.53}$ & $60.71$ & $64.19$ & $56.71$ \\
        DKT & $52.59_{\pm 7.37}$ & $53.20_{\pm 10.50}$ & $49.56_{\pm 4.73}$ & $59.90$ & $63.22$ & $55.18$ \\
        DKVMN & $47.22_{\pm 9.12}$ & $46.81_{\pm 8.04}$ & $48.80_{\pm 12.37}$ & $57.53$ & $60.42$ & $52.45$ \\
        AKT & $49.95_{\pm 4.74}$ & $51.40_{\pm 7.34}$ & $50.78_{\pm 11.04}$ & $62.07$ & $63.31$ & $57.14$ \\
        SAINT & $49.17_{\pm 4.88}$ & $47.81_{\pm 3.89}$ & $50.38_{\pm 6.68}$ & $55.97$ & $60.10$ & $54.24$ \\
        simpleKT & $51.25_{\pm 4.05}$ & $51.25_{\pm 5.60}$ & $49.82_{\pm 10.45}$ & $59.24$ & $63.83$ & $53.85$ \\
        \hline
        DKT-Sem & $\underline{56.83}_{\pm 9.75}$ & $\underline{61.82}_{\pm 6.56}$ & $\mathbf{63.27}_{\pm 9.57}$ & $\underline{62.17}$ & $\underline{66.18}$ & $\underline{59.30}$ \\
        LLMKT & $\mathbf{58.01}_{\pm 5.40}$ & $\mathbf{65.79}_{\pm 4.44}$ & $\underline{60.72}_{\pm 10.77}$ & $\mathbf{68.41}$ & $\mathbf{76.71}$ & $\mathbf{62.21}$ \\
        \bottomrule
    \end{tabular}
    \caption{Student dialogue turn correctness prediction performance (averaged with standard deviation shown when applicable) for all KT methods, with best numbers in \textbf{bold} and second-best \underline{underlined}. LLMKT outperforms all other methods on both datasets.}
    \label{tab:main_results}
\end{table*}

Table~\ref{tab:main_results} shows the quantitative performance of all KT methods on both datasets. We see that LLMKT significantly outperforms all existing KT methods and generally outperforms DKT-Sem on both datasets, confirming its effectiveness in the dialogue setting. This advantage over other KT methods is likely due to the use of a pre-trained LLM to capture the rich textual content in dialogues. As a result, the LLM can estimate student knowledge after fine-tuning on student turn response correctness labels.
We also see that DKT-Sem performs better than all existing KT methods, even though it largely uses the same model architecture as DKT. This result further emphasizes the benefit of utilizing the textual dialogue context in some way, even if the original model architecture of existing KT methods remains largely unchanged. 

On CoMTA, we see that all existing KT methods almost completely fail, unable to outperform the simple baseline of predicting the majority class label (which corresponds to an Acc.\ of $57.83_{\pm 5.08}$, AUC of $50.0_{\pm 0.00}$, and F1 of $58.87_{\pm 29.70}$). This result is expected since the size of the CoMTA dataset is small and neural network-based KT methods are data-hungry. Another important factor is that there are 164 unique KCs across only 623 labeled dialogue turn pairs, which is not enough for these methods to properly learn the KC embeddings or for BKT to learn the per-KC parameters. In contrast, LLMKT needs only minimal training data to reach acceptable accuracy, due to the underlying LLM being highly effective at understanding previous student dialogue turns. It can also leverage the textual content of the KC definitions to generalize across KCs, which is especially beneficial for ones that do not occur frequently. 

On MathDial, which has 21x more data than CoMTA, we see that existing KT methods perform significantly better (for reference, the majority baseline only gets an Acc.\ of $52.64$ since class labels are balanced). This result is expected and shows that they can perform the dialogueKT task when given enough training data.
While DKT-Sem still outperforms existing KT methods, it does so by a smaller margin than on CoMTA, indicating DKT-Sem has its greatest advantage when little training data is given.
However, LLMKT still maintains a large advantage over all other KT methods. This result suggests that even on large-scale datasets, LLM inference over prior dialogue history and textual content of KCs can still significantly improve KT performance, even when labeled student turn correctness data is abundant.

Finally, we see that performance of all KT methods is relatively low on the dialogueKT task, with a maximum of around $76\%$ AUC for student dialogue turn correctness prediction. This result is in stark contrast to results on the standard KT task, where state-of-the-art KT methods can surpass $80\%$ AUC for question response correctness prediction on some datasets. This observation suggests that dialogueKT is a challenging task, due to various reasons: First, dialogues can be relatively short, giving very little context for KT methods to leverage. For example, dialogues in CoMTA are cut short on purpose, with only 4-5 labeled turn pairs on average, to serve the goal in \cite{miller_dicerbo_2024} on evaluating how accurate LLMs are at math. Therefore, we call upon researchers who have access to dialogues with more turns to study these methods on their (possibly proprietary) data using our publicly available implementation. Second, student behavior in dialogues is more unpredictable compared to traditional KT, since there can be many diverse types of student responses to tutor turns. We believe that these challenges make dialogueKT a task worth studying in future work; it is likely that new KT methods need to be developed to improve performance on this task.

\subsection{Qualitative Analysis: Case Study}
\label{sec:qual}

\begin{table*}[t]
    \centering
    \small
    \begin{tabular}{lp{6cm}p{6.5cm}}
        \toprule
        Speaker & Turn Text & KCs and Predicted Masteries \\
        \midrule
        Tutor & Point X is at (-4, 6) and point Z is at (6, -2). Determine the coordinates of point Y on line segment XZ such that XY is 3 times as long as YZ. What do you think is the important information here? & 1) Use a pair of perpendicular number lines to define a coordinate system. \colorbox{yellow!40}{$0.4688$}\newline
        2) Apply the Pythagorean Theorem to find the distance between two points. \colorbox{yellow!40}{$0.4688$}\newline
        3) Recognize and represent proportional relationships between quantities. \colorbox{yellow!40}{$0.5622$}\newline
        4) Find the point on a line segment between two points that partitions the segment in a given ratio. \colorbox{green!40}{$0.6225$} \\
        Student & Firstly, I need to determine the distance between point X and Z. That would be the square root of 164, which is roughly 12.8 \color{green}\ding{51} & \\
        Tutor & Excellent beginning! You've calculated the total length of line segment XZ. Now, how can we use this information to determine the coordinates of point Y? Keep in mind, XY is 3 times as long as YZ. & 1) Recognize and represent proportional relationships between quantities. \colorbox{red!40}{$0.3208$}\newline
        2) Find the point on a line segment between two points that partitions the segment in a given ratio. \colorbox{red!40}{$0.3486$}\\
        Student & I would divide 12.8 by 3 total parts. \color{red}\ding{55} & \\
        \bottomrule
    \end{tabular}
    \caption{An example dialogue with student turn correctness labels, KC labels, and LLMKT's predicted mastery level for each KC. LLMKT accurately predicts student correctness at both turns by leveraging the difficulty of the tutor-posed questions.}
    \label{tab:qual_example}
\end{table*}

We now detail findings from our qualitative analysis, where we seek to i) identify challenges in the dialogueKT task, ii) understand why LLMKT has an advantage over existing KT methods, and most importantly, iii) show how LLMKT can be useful to tutors in practice. Therefore, we perform a case study using a dialogue in the test set of CoMTA.

We identify two primary challenges in the data that are unique to the dialogueKT task and are difficult for prior KT methods. First, the open-ended nature of dialogues means that student behavior is much less structured than in traditional assessment or practice settings. For example, when the tutor poses a question to the student, the student often i) does not answer directly but instead asks a clarifying question, ii) skips ahead to the final solution in the case of step-by-step problem solving, or iii) asks the tutor to give them a different problem. These student actions are inherently difficult to anticipate ahead of time since they define behavior beyond what can be captured by the KCs. As a result, there are many turns in these dialogues that are ``irrelevant'' to the KT task.

Second, since dialogues can be short (around 4 turn pairs on average in CoMTA), there can be little historical context to estimate student knowledge from. In many dialogues, students are proficient in one aspect of problem solving, responding to several questions correctly at first, but then struggle when the tutor moves on to a new KC. Without sufficiently long student history, it is difficult to estimate KC mastery levels for ones that are seen for the first time. This problem is partially due to existing datasets not linking students over multiple dialogues.

We find that, unsurprisingly, LLMKT's advantage over existing KT methods comes from leveraging nuanced details in the dialogue, rather than solely relying on previous correctness and KC labels. For the most part, changes in predicted mastery levels correlate with previous turn correctness; as a result, dialogues with many correct or incorrect turns in a row are easy to predict for all KT methods. LLMKT excels in cases when it relies on textual information, rather than correctness labels, to adjust its KC mastery estimates. Table~\ref{tab:qual_example} shows one such example: At the first turn, the tutor asks the student for relevant information regarding the problem, and the student responds correctly by applying the Pythagorean Theorem, which the model accurately predicts. However, after the tutor asks the student to find the coordinates of Y, the student incorrectly wants to divide the segment by 3 rather than 4, which the model also accurately predicts. The trained model seems to have estimated that the second tutor question is more difficult than the first, using the exact textual content of the tutor question, even though they have overlapping KC labels. This correct prediction is not possible without the question parsing ability enabled by using an LLM as the underlying KT model. 

We also found that LLMKT tends to predict low mastery when a KC is seen for the first time in a dialogue, possibly through observing that students do not react well to the tutor shifting to new KCs mid-way through the dialogue. This observation also suggests that LLMKT can effectively use dialogue context to detect when new KCs are introduced. Overall, these observations suggest that there are numerous advantages in exploiting the textual content in dialogues for the dialogueKT task, rather than just relying the correctness and KC labels; LLMKT seems to be an encouraging step in this direction for future KT methods. 

\subsection{``Learning Curve'' Analysis}

\begin{figure*}[t]
    \centering
    \includegraphics[height=3.3cm]{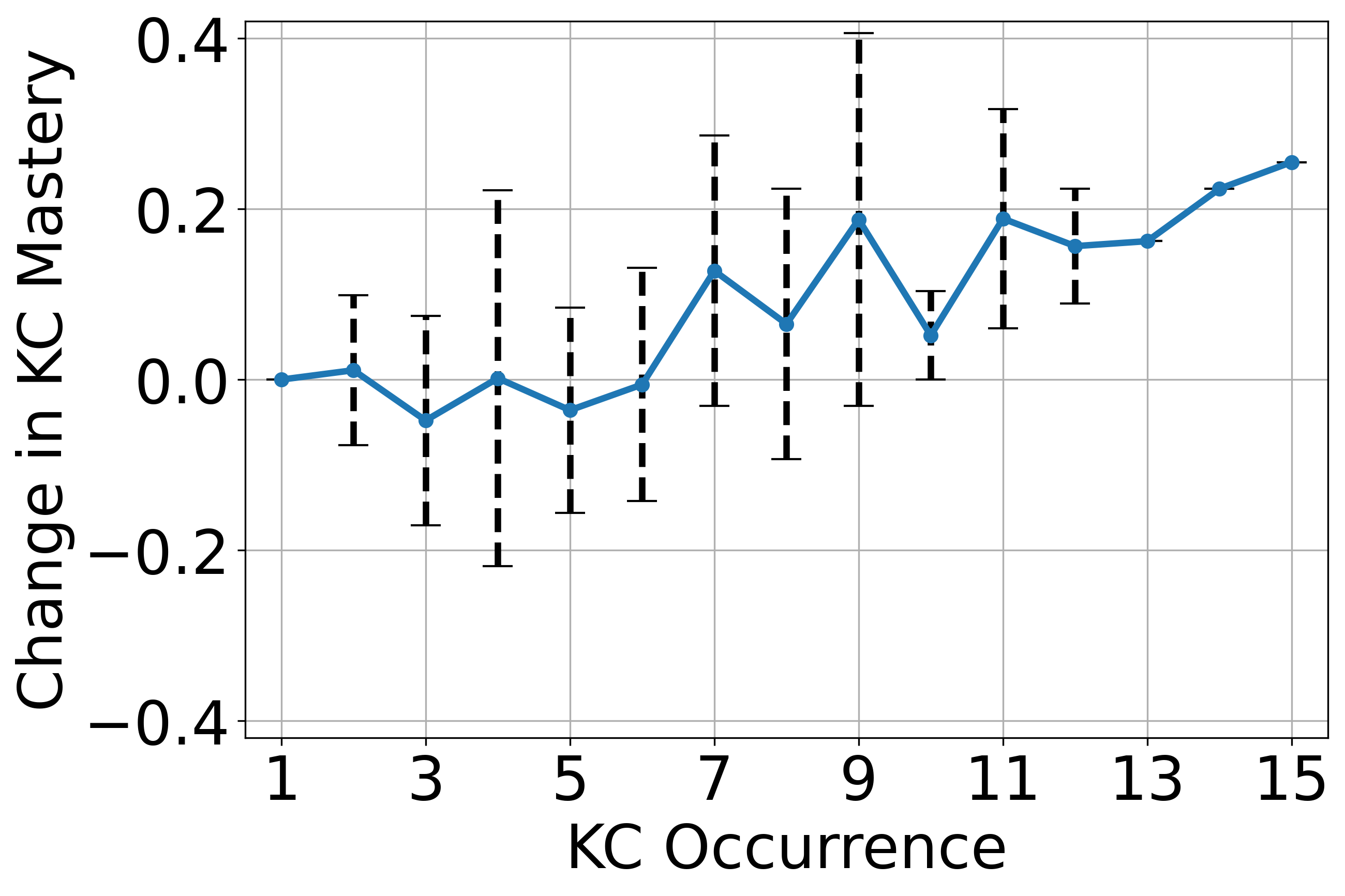}
    \includegraphics[height=3.3cm]{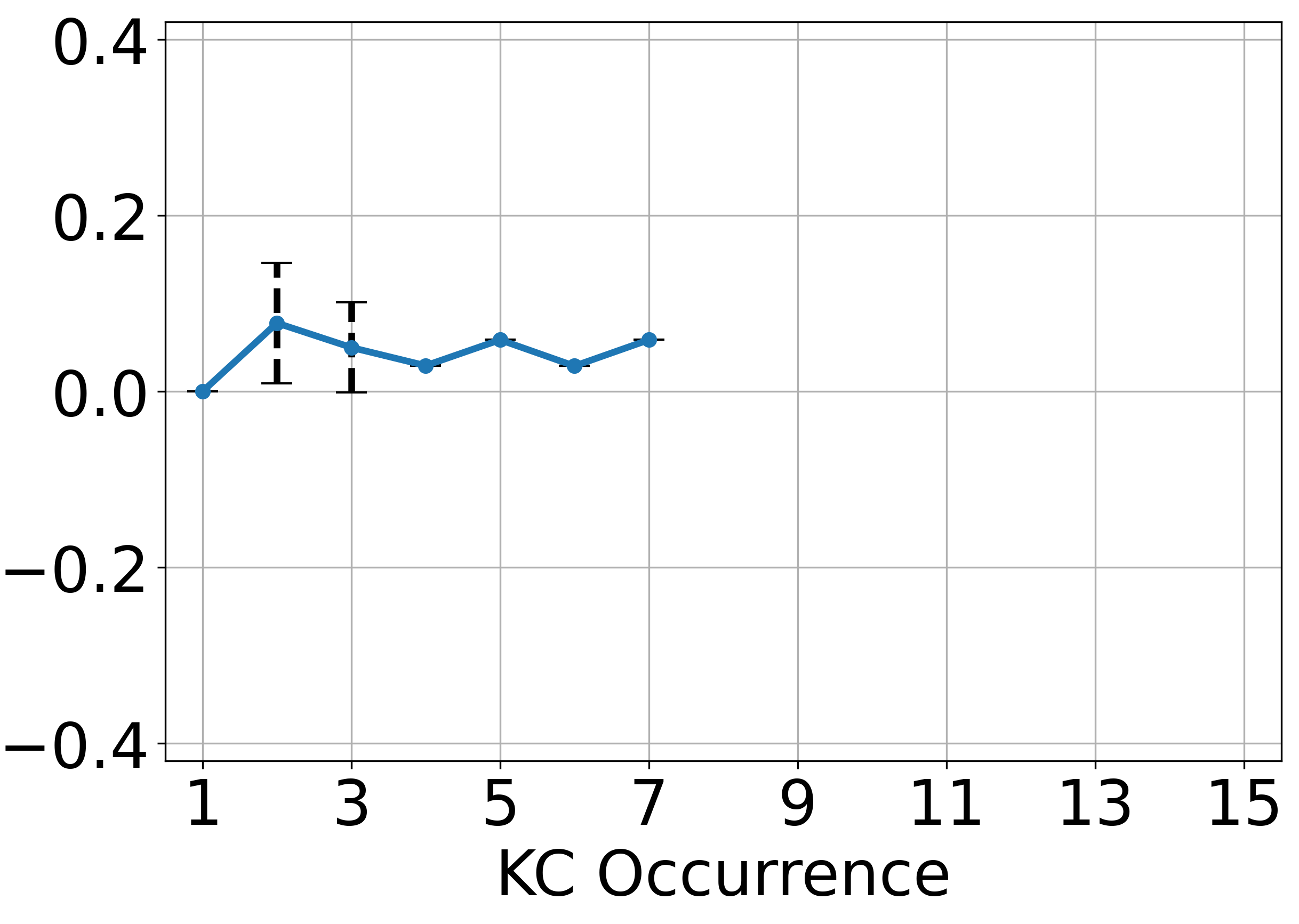}
    \includegraphics[height=3.3cm]{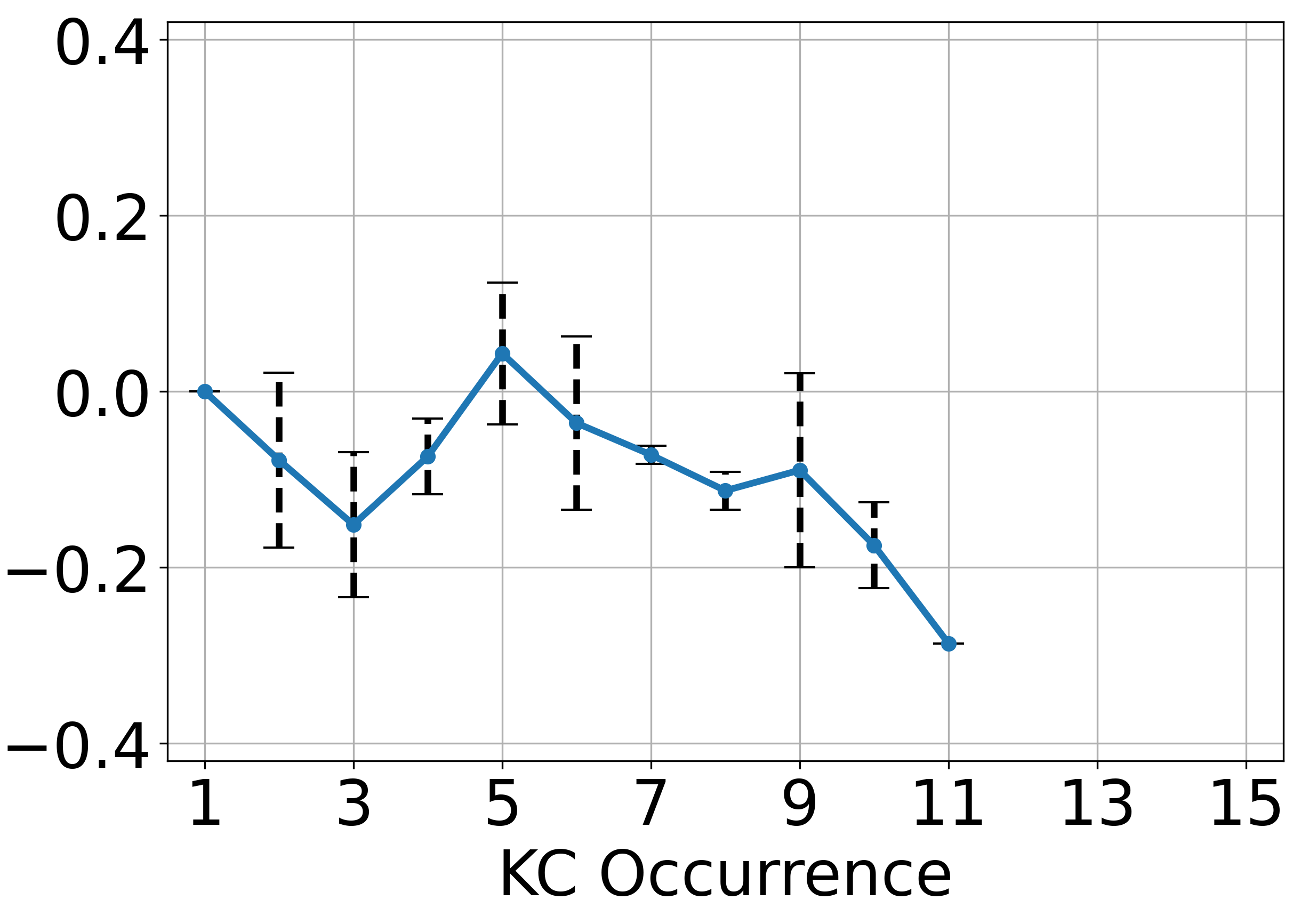}
    \caption{Knowledge change curves, i.e., estimated KC mastery level changes across dialogue turns, averaged over students for selected KCs in the CoMTA dataset. More common KCs in longer dialogues tend to show increasing mastery levels as the dialogue progresses, while less common KCs tend to show the opposite. KCs shown: ``Solve linear equations and inequalities in one variable,'' ``Understand that polynomials form a system analogous to the integers,'' and ``Apply the properties of operations to generate equivalent expressions.''}
    \label{fig:mastery-curves}
\end{figure*}

A common method for assessing the quality of KCs and KT methods is examining how well they match pedagogical or cognitive theory, and in particular, the power law of practice \cite{snoddy1926learning}: as students practice a particular skill, their mastery will increase rapidly at first and then more slowly over time. Following prior work \cite{kc-finder}, we visualize changes in predicted mastery levels at the KC-level, examine trends in the curves, and investigate whether there is evidence of the power law of practice. We use LLMKT's predicted masteries on CoMTA for all dialogues when they are in the test split of the cross-fold experiment. We use CoMTA instead of MathDial for this analysis since the latter uses LLMs to simulate students; it is unlikely for simulated students to behave like real ones, which we found in a preliminary analysis.

For a KC $c$, we first find each time that KC occurs in the data, use $c^{(i)}_j$ to denote its $j^\text{th}$ occurrence in the $i^\text{th}$ dialogue, and use $\hat{z}^{(i)}_j$ to denote the corresponding predicted KC mastery level. We then subtract $\hat{z}^{(i)}_1$ from each $\hat{z}^{(i)}_j$ to track the \textit{change} in predicted mastery from the first prediction, since different students may have different initial knowledge levels. We plot the average KC mastery level, i.e., $\textstyle \frac{1}{N} \sum_{i=1}^N \hat{z}^{(i)}_j - \hat{z}^{(i)}_1$, where $N$ is the number of dialogues that have at least $j$ turns that have $c$, and show the standard deviation using error bars.

We examine plots of the 15 most frequent KCs since less common KCs do not have enough data to show clear trends. Out of these, the results are mixed: 5 out of 15 KCs trend towards increasing mastery as the dialogues go on, 5 KCs trend towards decreasing mastery, and 5 KCs do not change significantly in mastery. Figure~\ref{fig:mastery-curves} shows a representative example of each case. KCs with an upward trend typically have few changes in their masteries for the first few turns, as well as relatively high variance. Additionally, these KCs are more prevalent in longer dialogues and correspond to the more commonly occurring KCs in the data. These patterns can be explained by two concurrent trends in the data: First, there are several long dialogues where a student struggles with a math topic at the beginning, getting some questions correct and others incorrect, but after some time becomes more proficient and consistently responds correctly; these dialogues contribute to the steady increase in KC mastery. Second, there are several short dialogues where the student either simply knows the topic and consistently responds correctly, or struggles with the topic and responds incorrectly; these dialogues contribute to the high variance and the minimal change in early turns.

On the contrary, KCs with a downward trend or no trend at all tend to be associated with shorter dialogues and occur in fewer dialogues. These dialogues tend to feature a student repeatedly trying to solve a problem without any success, with the dialogue often ending immediately after a correct solution is reached. KC mastery can also drop significantly at the beginning of the dialogue once the student starts to respond incorrectly. Since there are few dialogues to average over, these downward trends can dominate the knowledge change curves; the most common KC with a downward trend only appears in 8 dialogues, which is less than most KCs with upward trends. 

Overall, when we take into account the small amount of data and how dialogues in the CoMTA dataset are truncated, the knowledge change curve trends seem to mostly resemble the power law of practice; when there are sufficient turns in a dialogue, the student's predicted mastery increases before leveling out eventually. The main reason these trends are not more prevalent is likely due to these dialogues typically ending after only a few turns. This setting is different than the problem-solving setup mostly used in past KT research, where students practice a KC repeatedly by solving different problems, even after mostly mastering it. Again, we call upon researchers who have access to data with longer dialogues to conduct more extensive analyses. 

\section{Annotation Quality Experiments}
\label{sec:annotation-experiments}

\subsection{Human Evaluation}

To quantify GPT-4o's accuracy on the correctness and KC labeling tasks, we conduct an IRB-approved human evaluation using expert human annotators. We hired 3 former math teachers at \$40 US/hour, each rating the same set of 30 randomly selected dialogues from CoMTA with GPT-4o's labels, totaling 166 turn pairs. For each turn in each dialogue, we ask annotators to score GPT-4o's label i) on correctness, as 1 if they think the label is accurate and 0 otherwise, and ii) on the standards selected by GPT-4o on a 1-to-4 scale with 4 as the highest. We provide the exact instructions given to annotators in the online Supplementary Material\footnote{\url{https://osf.io/wrkej?view_only=f32472a6560649e692a4e2ba603ef52c}}. For both correctness and KCs, we report the average human-assigned \textbf{Score} and two inter-rater reliability (IRR) metrics: i) the portion of times that the labels of all 3 annotators are exactly the same (\textbf{Overlap}) and ii) the Krippendorff's $\boldsymbol{\alpha}$ coefficient \cite{castro-2017-fast-krippendorff}. We also compute the accuracy of correctness predictions at the final turn using CoMTA's ground truth labels for both GPT-4o and our annotators.

\begin{table*}[]
    \centering
    \begin{tabular}{lccccc}
        \toprule
        Label & Score & Overlap & $\alpha$ & GPT-4o Final Turn Acc. & Human Final Turn Acc. \\
        \midrule
        Correctness & 0.9317 / 1 & 0.8434 & 0.1809 & 75.82 (83.33) & 85.56 \\
        KCs & 3.2831 / 4 & 0.3494 & 0.4383 & -- & -- \\
        \bottomrule
    \end{tabular}
    \caption{Scores and IRR metrics for the human evaluation. Additionally, final turn correctness accuracy for GPT-4o (on the full dataset followed by only on the human evaluation dataset) and the human annotators.}
    \label{tab:anno_results}
\end{table*}

Table~\ref{tab:anno_results} shows the human evaluation results. We see that our annotators give GPT-4o very high scores for correctness and moderate-to-high scores for KCs, indicating that GPT-4o is able to competently perform these tasks. IRR is volatile, with high overlap and low $\alpha$ on correctness, and low overlap but moderate $\alpha$ on KCs. However, the low $\alpha$ on correctness is expected given the high class imbalance of the data (more than 90\% of labels are 1), and the low overlap on KCs is expected since there are 4 classes and exact ratings can be subjective. While GPT-4o's accuracy on final turn correctness is lower than the correctness score might suggest, this is expected since the final turns are much harder to compute correctness on without subsequent tutor turns to help indicate the label. 
As reference, the CoMTA tech report \cite{miller_dicerbo_2024} reports final turn correctness evaluation with GPT-4o at similar accuracy (78.3). 
Finally, given that the average human annotator's final turn correctness accuracy is only slightly higher than GPT-4o's, we can conclude that the correctness labeling task is a challenging one where GPT-4o achieves close to human-level performance.

\subsection{Error Analysis}

To understand when GPT-4o makes errors in labeling correctness and KCs, we qualitatively analyze its annotations.
When labeling correctness, most of GPT-4o's errors are on the final turn, especially in questions that require some numerical calculations to verify correctness, such as when the student provides the numerical result of a sum or product calculation. This is a known limitation in GPT-4/4o \cite{scarlatos2024feedback} and can possibly be improved through external tools such as calculators. 
The other common error tends to be when GPT-4o should label a turn as ``na'' but does not, e.g., when the tutor asks the student whether they know how to use a function on their calculator. GPT-4o also sometimes labels a turn as correct when the tutor asks a student if they know a concept and the student indicates they do not, showing that GPT-4o can be confused by the tutor's intentions.

When labeling KCs, GPT-4o's primary error is not assigning sufficient standards to a turn to describe all the required KCs. This error is mostly due to not selecting enough relevant domains and clusters early in the annotation process, especially for ones with broad or abstract descriptions. As a result, there are not enough related standards to select from to use as KCs. On the other hand, GPT-4o occasionally assigns too many KCs to a turn, selecting all standards that are related to the topic but not necessarily only ones relevant to the current question. These errors can likely be resolved with more sophisticated prompting, potentially running the annotation process multiple times through sampling and taking a majority vote. 
We leave the exploration of such methods that require more API calls for future work. Finally, there are several dialogues not covered by Common Core standards; there is little that GPT-4o can do other than select a somewhat related standard. This observation suggests that it may be necessary to augment the Common Core standards with other sources to have a more complete set of candidate KCs, or defer to humans for labeling in these cases.

\section{Discussion, Conclusions, and Future Work}

In this paper, we introduced the task of dialogueKT with the goal of estimating student knowledge from open-ended tutor-student dialogues. We used GPT-4o to automatically annotate student response correctness and relevant KCs in the dialogue turns, and showed via an expert human evaluation that these annotations are largely accurate. We introduced a new LLM-based KT method that leverages the textual content in dialogues, and showed that LLMKT i) significantly outperforms existing methods on dialogueKT and ii) works with limited training data. We also showed that our models use the textual content in dialogues to capture student knowledge in meaningful ways.
There are numerous avenues for future work. First, we can jointly model tutor moves and student knowledge states to analyze what types of tutor moves are effective at improving student knowledge. Second, in addition to student knowledge, we also need to account for student affect dynamics in dialogues, possibly with the help of other data modalities such as audio and video. Third, the scope of dialogueKT can be expanded upon: in this work, we only focus on math dialogues between a single student and tutor, but future work can extend to other subjects and dialogues involving multiple students. Fourth, there are many potential applications of dialogueKT that can be explored. Future work should investigate whether estimated knowledge states are useful for personalization, providing teacher support, or can even be used to train AI-based tutors by powering simulated student agents. Overall, we hope that this initial investigation can broaden the scope of student modeling to conversational settings, which have become of significant interest lately.

\section{Acknowledgments}
The authors would like to thank Kristen Dicerbo and the Khan Academy team for guidelines on how to use the CoMTA dataset and releasing it. We would also like to thank Nigel Fernandez, Zichao Wang, and Jacob Whitehill for helpful discussions around this work. This work is partially supported by Schmidt Futures under the learning engineering virtual institute (LEVI) initiative and the NSF under grants 2118706, 2202506, 2237676, and 2341948. 

\bibliographystyle{plain}
\bibliography{references}

\begin{thebibliography}{10}

\bibitem{abdelshiheed2024aligning}
Mark Abdelshiheed, Jennifer K.~Jacobs, and Sidney K.~D'Mello.
\newblock Aligning tutor discourse supporting rigorous thinking with tutee content mastery for predicting math achievement.
\newblock In {\em Artificial Intelligence in Education}, pages 150--164, Cham, 2024. Springer Nature Switzerland.

\bibitem{pybkt}
Anirudhan Badrinath, Frederic Wang, and Zachary Pardos.
\newblock pybkt: An accessible python library of bayesian knowledge tracing models.
\newblock In {\em Proceedings of the 14th International Conference on Educational Data Mining}, pages 468--474, 2021.

\bibitem{barnes2005q}
Tiffany Barnes.
\newblock The q-matrix method: Mining student response data for knowledge.
\newblock In {\em American association for artificial intelligence 2005 educational data mining workshop}, pages 1--8. AAAI Press, Pittsburgh, PA, USA, 2005.

\bibitem{cai2018impact}
Zhiqiang Cai, Arthur~C Graesser, Leah Windsor, Qinyu Cheng, David~W Shaffer, and Xiangen Hu.
\newblock Impact of corpus size and dimensionality of lsa spaces from wikipedia articles on autotutor answer evaluation.
\newblock {\em Journal of educational data mining}, 2018.

\bibitem{livehint}
{Carnegie Learning}.
\newblock Livehint overview.
\newblock Online: \url{https://support.carnegielearning.com/help-center/math/livehint/article/livehint-overview/}, 2024.

\bibitem{castro-2017-fast-krippendorff}
Santiago Castro.
\newblock Fast {K}rippendorff: Fast computation of {K}rippendorff's alpha agreement measure.
\newblock \url{https://github.com/pln-fing-udelar/fast-krippendorff}, 2017.

\bibitem{chang2018dialogue}
Maria Chang, Matthew Ventura, Jae-wook Ahn, Peter Foltz, Tengfei Ma, Tejas~I Dhamecha, Smit Marvaniya, Patrick Watson, Cassius D’helon, Amy Wetzel, et~al.
\newblock Dialogue-based tutoring at scale: Design and challenges.
\newblock In {\em Practitioner and Industrial Track Proceedings of the 13th International Conference of the Learning Sciences}, 2018.

\bibitem{das3h}
Beno{\^\i}t Choffin, Fabrice Popineau, Yolaine Bourda, and Jill-J{\^e}nn Vie.
\newblock {DAS3H}: {M}odeling student learning and forgetting for pptimally scheduling distributed practice of skills.
\newblock In {\em Proc. Int. Conf. Educ. Data Mining}, pages 29--38, 2019.

\bibitem{saint}
Youngduck Choi, Youngnam Lee, Junghyun Cho, Jineon Baek, Byungsoo Kim, Yeongmin Cha, Dongmin Shin, Chan Bae, and Jaewe Heo.
\newblock Towards an appropriate query, key, and value computation for knowledge tracing.
\newblock In {\em Proceedings of the Seventh ACM Conference on Learning @ Scale}, L@S '20, page 341–344, New York, NY, USA, 2020. Association for Computing Machinery.

\bibitem{commoncore}
{Common Core State Standards Initiative}.
\newblock Mathematics standards.
\newblock Online: \url{https://www.thecorestandards.org/Math/}, 2024.

\bibitem{kt}
Albert Corbett and John Anderson.
\newblock Knowledge tracing: {M}odeling the acquisition of procedural knowledge.
\newblock {\em User Model. User-adapted Interact.}, 4(4):253--278, Dec. 1994.

\bibitem{croteau2004algebra}
Ethan~A Croteau, Neil~T Heffernan, and Kenneth~R Koedinger.
\newblock Why are algebra word problems difficult? using tutorial log files and the power law of learning to select the best fitting cognitive model.
\newblock In {\em Intelligent Tutoring Systems: 7th International Conference, ITS 2004, Macei{\'o}, Alagoas, Brazil, August 30-September 3, 2004. Proceedings 7}, pages 240--250. Springer, 2004.

\bibitem{cui2023adaptive}
Peng Cui and Mrinmaya Sachan.
\newblock Adaptive and personalized exercise generation for online language learning.
\newblock In {\em Proceedings of the 61st Annual Meeting of the Association for Computational Linguistics (Volume 1: Long Papers)}, pages 10184--10198, 2023.

\bibitem{ncte}
Dorottya Demszky and Heather Hill.
\newblock The ncte transcripts: A dataset of elementary math classroom transcripts.
\newblock In {\em Proceedings of the 18th Workshop on Innovative Use of NLP for Building Educational Applications (BEA 2023)}, pages 528--538, 2023.

\bibitem{demszky2021measuring}
Dorottya Demszky, Jing Liu, Zid Mancenido, Julie Cohen, Heather Hill, Dan Jurafsky, and Tatsunori~B Hashimoto.
\newblock Measuring conversational uptake: A case study on student-teacher interactions.
\newblock In {\em Proceedings of the 59th Annual Meeting of the Association for Computational Linguistics and the 11th International Joint Conference on Natural Language Processing (Volume 1: Long Papers)}, pages 1638--1653, 2021.

\bibitem{fernandez2024divert}
Nigel Fernandez, Alexander Scarlatos, Wanyong Feng, Simon Woodhead, and Andrew Lan.
\newblock {D}i{VERT}: Distractor generation with variational errors represented as text for math multiple-choice questions.
\newblock In {\em Proceedings of the 2024 Conference on Empirical Methods in Natural Language Processing}, pages 9063--9081, Miami, Florida, USA, November 2024. Association for Computational Linguistics.

\bibitem{akt}
Aritra Ghosh, Neil Heffernan, and Andrew~S Lan.
\newblock Context-aware attentive knowledge tracing.
\newblock In {\em Proc. ACM SIGKDD}, pages 2330--2339, 2020.

\bibitem{glickman2012example}
Mark~E Glickman.
\newblock Example of the glicko-2 system.
\newblock {\em Boston University}, 28, 2012.

\bibitem{learnlm}
{Google}.
\newblock How generative ai expands curiosity and understanding with learnlm.
\newblock Online: \url{https://blog.google/outreach-initiatives/education/google-learnlm-gemini-generative-ai/}, 2024.

\bibitem{llama31}
Aaron Grattafiori and Others.
\newblock The llama 3 herd of models, 2024.

\bibitem{he2021quizzing}
Joy He-Yueya and Adish Singla.
\newblock Quizzing policy using reinforcement learning for inferring the student knowledge state.
\newblock {\em International Educational Data Mining Society}, 2021.

\bibitem{hu2022lora}
Edward~J Hu, yelong shen, Phillip Wallis, Zeyuan Allen-Zhu, Yuanzhi Li, Shean Wang, Lu~Wang, and Weizhu Chen.
\newblock Lo{RA}: Low-rank adaptation of large language models.
\newblock In {\em International Conference on Learning Representations}, 2022.

\bibitem{jordan2001tools}
Pamela Jordan, Carolyn~Penstein Ros{\'e}, and Kurt VanLehn.
\newblock Tools for authoring tutorial dialogue knowledge.
\newblock In {\em Proceedings of AI in Education 2001 Conference}. IOS Press Amsterdam, 2001.

\bibitem{jung2024clstcoldstartmitigationknowledge}
Heeseok Jung, Jaesang Yoo, Yohaan Yoon, and Yeonju Jang.
\newblock Clst: Cold-start mitigation in knowledge tracing by aligning a generative language model as a students' knowledge tracer, 2024.

\bibitem{kaser2024simulated}
Tanja K{\"a}ser and Giora Alexandron.
\newblock Simulated learners in educational technology: A systematic literature review and a turing-like test.
\newblock {\em International Journal of Artificial Intelligence in Education}, 34(2):545--585, 2024.

\bibitem{khanmigo}
{Khan Academy}.
\newblock Supercharge your teaching experience with khanmigo.
\newblock Online: \url{https://www.khanmigo.ai/}, 2023.

\bibitem{lee2024languagemodelknowledgetracing}
Unggi Lee, Jiyeong Bae, Dohee Kim, Sookbun Lee, Jaekwon Park, Taekyung Ahn, Gunho Lee, Damji Stratton, and Hyeoncheol Kim.
\newblock Language model can do knowledge tracing: Simple but effective method to integrate language model and knowledge tracing task, 2024.

\bibitem{li2024knowledgetaggingmathquestions}
Hang Li, Tianlong Xu, Jiliang Tang, and Qingsong Wen.
\newblock Knowledge tagging system on math questions via llms with flexible demonstration retriever, 2024.

\bibitem{li2024explainable}
Haoxuan Li, Jifan Yu, Yuanxin Ouyang, Zhuang Liu, Wenge Rong, Juanzi Li, and Zhang Xiong.
\newblock Explainable few-shot knowledge tracing.
\newblock {\em arXiv preprint arXiv:2405.14391}, 2024.

\bibitem{okt}
Naiming Liu, Zichao Wang, Richard Baraniuk, and Andrew Lan.
\newblock Open-ended knowledge tracing for computer science education.
\newblock In {\em Proceedings of the 2022 Conference on Empirical Methods in Natural Language Processing}, 2022.

\bibitem{eernna}
Qi~Liu, Zhenya Huang, Yu~Yin, Enhong Chen, Hui Xiong, Yu~Su, and Guoping Hu.
\newblock Ekt: Exercise-aware knowledge tracing for student performance prediction.
\newblock {\em IEEE Trans. Knowl. Data Eng.}, 33(1):100--115, 2019.

\bibitem{liu2023simplekt}
Zitao Liu, Qiongqiong Liu, Jiahao Chen, Shuyan Huang, and Weiqi Luo.
\newblock simple{KT}: A simple but tough-to-beat baseline for knowledge tracing.
\newblock In {\em The Eleventh International Conference on Learning Representations}, 2023.

\bibitem{liupykt2022}
Zitao Liu, Qiongqiong Liu, Jiahao Chen, Shuyan Huang, Jiliang Tang, and Weiqi Luo.
\newblock pykt: A python library to benchmark deep learning based knowledge tracing models.
\newblock In {\em Thirty-sixth Conference on Neural Information Processing Systems Datasets and Benchmarks Track}, 2022.

\bibitem{atc}
Li~Lucy, Tal August, Rose~E Wang, Luca Soldaini, Courtney Allison, and Kyle Lo.
\newblock {M}ath{F}ish: Evaluating language model math reasoning via grounding in educational curricula.
\newblock In {\em Findings of the Association for Computational Linguistics: EMNLP 2024}, pages 5644--5673, Miami, Florida, USA, November 2024. Association for Computational Linguistics.

\bibitem{ma2024debugging}
Qianou Ma, Hua Shen, Kenneth Koedinger, and Sherry~Tongshuang Wu.
\newblock How to teach programming in the ai era? using llms as a teachable agent for debugging.
\newblock In {\em Artificial Intelligence in Education}, pages 265--279, Cham, 2024. Springer Nature Switzerland.

\bibitem{macina-etal-2023-mathdial}
Jakub Macina, Nico Daheim, Sankalan Chowdhury, Tanmay Sinha, Manu Kapur, Iryna Gurevych, and Mrinmaya Sachan.
\newblock {M}ath{D}ial: A dialogue tutoring dataset with rich pedagogical properties grounded in math reasoning problems.
\newblock In Houda Bouamor, Juan Pino, and Kalika Bali, editors, {\em Findings of the Association for Computational Linguistics: EMNLP 2023}, pages 5602--5621, Singapore, December 2023. Association for Computational Linguistics.

\bibitem{maier2021challenges}
Cristina Maier, Ryan Baker, and Steve Stalzer.
\newblock Challenges to applying performance factor analysis to existing learning systems.
\newblock In {\em International Conference on Computers in Education}, 2021.

\bibitem{makatchev2007combining}
Maxim Makatchev and Kurt VanLehn.
\newblock Combining bayesian networks and formal reasoning for semantic classification of student utterances.
\newblock In {\em Proceedings of the 2007 conference on Artificial Intelligence in Education: Building Technology Rich Learning Contexts That Work}, pages 307--314, 2007.

\bibitem{miller_dicerbo_2024}
Pepper Miller and Kristen DiCerbo.
\newblock Llm based math tutoring: Challenges and dataset, Jul 2024.

\bibitem{metaat}
Andre Nickow, Philip Oreopoulos, and Vincent Quan.
\newblock The impressive effects of tutoring on prek-12 learning: A systematic review and meta-analysis of the experimental evidence.
\newblock Working Paper 27476, National Bureau of Economic Research, July 2020.

\bibitem{nye2014autotutor}
Benjamin~D Nye, Arthur~C Graesser, and Xiangen Hu.
\newblock Autotutor and family: A review of 17 years of natural language tutoring.
\newblock {\em International Journal of Artificial Intelligence in Education}, 24:427--469, 2014.

\bibitem{gpt-4o}
OpenAI.
\newblock Hello gpt-4o.
\newblock Online: \url{https://openai.com/index/hello-gpt-4o/}, May 2024.

\bibitem{sakt}
Shalini Pandey and George Karypis.
\newblock A self attentive model for knowledge tracing.
\newblock In {\em Proc. Int. Conf. Educ. Data Mining}, pages 384--389, July 2019.

\bibitem{pfa}
Philip Pavlik~Jr, Hao Cen, and Kenneth Koedinger.
\newblock Performance factors analysis--{A} new alternative to knowledge tracing.
\newblock In {\em Proc. Int. Conf. Artif. Intell. Educ.}, 2009.

\bibitem{pelanek2017bayesian}
Radek Pel{\'a}nek.
\newblock Bayesian knowledge tracing, logistic models, and beyond: an overview of learner modeling techniques.
\newblock {\em User modeling and user-adapted interaction}, 27:313--350, 2017.

\bibitem{dkt}
Chris Piech, Jonathan Bassen, Jonathan Huang, Surya Ganguli, Mehran Sahami, Leonidas~J Guibas, and Jascha Sohl-Dickstein.
\newblock Deep knowledge tracing.
\newblock In {\em Proc. NeurIPS}, pages 505--513, 2015.

\bibitem{prihar2023comparing}
{Prihar, Ethan and Others}.
\newblock Comparing different approaches to generating mathematics explanations using large language models.
\newblock In {\em International Conference on Artificial Intelligence in Education}, pages 290--295, 2023.

\bibitem{rafferty2016faster}
Anna~N Rafferty, Emma Brunskill, Thomas~L Griffiths, and Patrick Shafto.
\newblock Faster teaching via pomdp planning.
\newblock {\em Cognitive science}, 40(6):1290--1332, 2016.

\bibitem{sbert}
Nils Reimers and Iryna Gurevych.
\newblock Sentence-bert: Sentence embeddings using siamese bert-networks.
\newblock In {\em Proceedings of the 2019 Conference on Empirical Methods in Natural Language Processing}. Association for Computational Linguistics, 11 2019.

\bibitem{scarlatos2024feedback}
Alexander Scarlatos, Digory Smith, Simon Woodhead, and Andrew Lan.
\newblock Improving the validity of automatically generated feedback via reinforcement learning.
\newblock In {\em Artificial Intelligence in Education}, pages 280--294, Cham, 2024. Springer Nature Switzerland.

\bibitem{shen2021classifying}
Jia~Tracy Shen, Michiharu Yamashita, Ethan Prihar, Neil Heffernan, Xintao Wu, Sean McGrew, and Dongwon Lee.
\newblock Classifying math knowledge components via task-adaptive pre-trained bert.
\newblock In {\em Artificial Intelligence in Education: 22nd International Conference, AIED 2021, Utrecht, The Netherlands, June 14--18, 2021, Proceedings, Part I 22}, pages 408--419. Springer, 2021.

\bibitem{kc-finder}
Yang Shi, Robin Schmucker, Min Chi, Tiffany Barnes, and Thomas Price.
\newblock Kc-finder: Automated knowledge component discovery for programming problems.
\newblock {\em International Educational Data Mining Society}, 2023.

\bibitem{saint+}
Dongmin Shin, Yugeun Shim, Hangyeol Yu, Seewoo Lee, Byungsoo Kim, and Youngduck Choi.
\newblock Saint+: Integrating temporal features for ednet correctness prediction.
\newblock In {\em 11th Int. Learn. Analytics Knowl. Conf.}, pages 490--496, 2021.

\bibitem{snoddy1926learning}
George~S Snoddy.
\newblock Learning and stability: a psychophysiological analysis of a case of motor learning with clinical applications.
\newblock {\em Journal of Applied Psychology}, 10(1):1, 1926.

\bibitem{srivastava2021question}
Megha Srivastava and Noah Goodman.
\newblock Question generation for adaptive education.
\newblock In {\em Proc. Annual Meeting of the Association for Computational Linguistics and the International Joint Conference on Natural Language Processing (Volume 2: Short Papers)}, pages 692--701, 2021.

\bibitem{suresh-etal-2022-talkmoves}
Abhijit Suresh, Jennifer Jacobs, Charis Harty, Margaret Perkoff, James~H. Martin, and Tamara Sumner.
\newblock The {T}alk{M}oves dataset: K-12 mathematics lesson transcripts annotated for teacher and student discursive moves.
\newblock In {\em Proceedings of the Thirteenth Language Resources and Evaluation Conference}, pages 4654--4662, June 2022.

\bibitem{suresh2022fine}
Abhijit Suresh, Jennifer Jacobs, Margaret Perkoff, James~H Martin, and Tamara Sumner.
\newblock Fine-tuning transformers with additional context to classify discursive moves in mathematics classrooms.
\newblock In {\em Proc. Workshop on Innovative Use of NLP for Building Educational Applications}, 2022.

\bibitem{vanlehn2002architecture}
Kurt VanLehn, Pamela~W Jordan, Carolyn~P Ros{\'e}, Dumisizwe Bhembe, Michael B{\"o}ttner, Andy Gaydos, Maxim Makatchev, Umarani Pappuswamy, Michael Ringenberg, Antonio Roque, et~al.
\newblock The architecture of why2-atlas: A coach for qualitative physics essay writing.
\newblock In {\em Intelligent Tutoring Systems: 6th International Conference, ITS 2002 Biarritz, France and San Sebastian, Spain, June 2--7, 2002 Proceedings 6}, pages 158--167. Springer, 2002.

\bibitem{wang2024bridging}
Rose Wang, Qingyang Zhang, Carly Robinson, Susanna Loeb, and Dorottya Demszky.
\newblock Bridging the novice-expert gap via models of decision-making: A case study on remediating math mistakes.
\newblock In {\em Proceedings of the 2024 Conference of the North American Chapter of the Association for Computational Linguistics: Human Language Technologies (Volume 1: Long Papers)}, pages 2174--2199, 2024.

\bibitem{wolf-etal-2020-transformers}
{Wolf, Thomas and Others}.
\newblock Transformers: State-of-the-art natural language processing.
\newblock In {\em Proceedings of the 2020 Conference on Empirical Methods in Natural Language Processing: System Demonstrations}, pages 38--45, October 2020.

\bibitem{yang2024content}
Kaiqi Yang, Yucheng Chu, Taylor Darwin, Ahreum Han, Hang Li, Hongzhi Wen, Yasemin Copur-Gencturk, Jiliang Tang, and Hui Liu.
\newblock Content knowledge identification with multi-agent large language models (llms).
\newblock In {\em Artificial Intelligence in Education}, pages 284--292, 2024.

\bibitem{gikt}
Yang Yang, Jian Shen, Yanru Qu, Yunfei Liu, Kerong Wang, Yaoming Zhu, Weinan Zhang, and Yong Yu.
\newblock Gikt: A graph-based interaction model for knowledge tracing.
\newblock In {\em Proc. Joint Eur. Conf. Mach. Learn. Knowl. Discovery Databases}, 2020.

\bibitem{zhan2024knowledge}
Bojun Zhan, Teng Guo, Xueyi Li, Mingliang Hou, Qianru Liang, Boyu Gao, Weiqi Luo, and Zitao Liu.
\newblock Knowledge tracing as language processing: A large-scale autoregressive paradigm.
\newblock In {\em Artificial Intelligence in Education}, pages 177--191, Cham, 2024. Springer Nature Switzerland.

\bibitem{dkvmn}
Jiani Zhang, Xingjian Shi, Irwin King, and Dit-Yan Yeung.
\newblock Dynamic key-value memory networks for knowledge tracing.
\newblock In {\em Proc. Int. Conf. World Wide Web}, pages 765--774, Apr. 2017.

\end{thebibliography}

\end{document}